\documentclass{article}

\usepackage{amssymb}
\usepackage{amsmath}
\usepackage{PRIMEarxiv}
\usepackage{subcaption}
\usepackage{multirow}
\usepackage[utf8]{inputenc} 
\usepackage[T1]{fontenc}   
\usepackage{hyperref}      
\usepackage{xurl} 
\usepackage{booktabs}      
\usepackage{amsfonts}      
\usepackage{nicefrac}       
\usepackage{microtype}     
\usepackage{lipsum}
\usepackage{fancyhdr}      
\usepackage{graphicx}

\title{Vision-Language Model for Accurate Crater Detection

}

\author{
\textbf{Patrick Bauer} \\
University of Technology of Troyes \\
Hochschule Darmstadt \\
\texttt{patrick.bauer@utt.fr}
\And
\textbf{Marius Schwinning} \\
GMV for European Space Agency\\
\texttt{marius.schwinning@esa.int}
\And
\textbf{Florian Renk} \\
European Space Agency\\
\texttt{florian.renk@esa.int}
\And
\textbf{Andreas Weinmann} \\
Technische Hochschule Würzburg-Schweinfurt \\
Hochschule Darmstadt\\
\texttt{andreas.weinmann@thws.de}
\And
\textbf{Hichem Snoussi} \\
University of Technology of Troyes \\
\texttt{hichem.snoussi@utt.fr}
}

\begin{document}

\noindent{\footnotesize\copyright~2026 IEEE. Personal use of this material is permitted.
Permission from IEEE must be obtained for all other uses, in any current or future media,
including reprinting/republishing this material for advertising or promotional purposes,
creating new collective works, for resale or redistribution to servers or lists, or reuse
of any copyrighted component of this work in other works.\par}
\vspace{1em}

\vspace{0.5em}

\noindent{\footnotesize
Accepted for publication in \textit{IEEE Transactions on Aerospace and Electronic Systems}.
DOI: \href{https://doi.org/10.1109/TAES.2026.3725640}{10.1109/TAES.2026.3725640}\par}

\vspace{1em}
\maketitle

\begin{abstract}
The European Space Agency (ESA), driven by its ambitions on planned lunar missions with the Argonaut lander, has a profound interest in reliable crater detection, since craters pose a risk to safe lunar landings. This task is usually addressed with automated crater detection algorithms (CDA) based on deep learning techniques. It is non-trivial due to the vast amount of craters of various sizes and shapes, as well as challenging conditions such as varying illumination and rugged terrain. Therefore, we propose a deep-learning CDA based on  the OWLv2 model, which is built on a Vision Transformer, that has proven highly effective in various computer vision tasks. For fine-tuning, we utilize a manually labeled dataset fom the IMPACT project, that provides crater annotations on high-resolution Lunar Reconnaissance Orbiter Camera Calibrated Data Record images. We insert trainable parameters using a parameter-efficient fine-tuning strategy with Low-Rank Adaptation, and optimize a combined loss function consisting of Complete Intersection over Union (CIoU) for localization and a contrastive loss for classification. We achieve satisfactory visual results, along with a maximum recall of $92.6\%$ and a maximum precision of $71.4\%$ on a test dataset from IMPACT. Our method achieves reliable crater detection across challenging lunar imaging conditions, paving the way for robust crater analysis in future lunar exploration. 
\end{abstract}

\keywords{Crater Detection \and Vision Language Models \and OWLv2 model}

\section{INTRODUCTION}
Craters are among the most prominent features on the lunar surface, and their accurate detection is of critical importance to the European Space Agency (ESA) due to its direct impact on the success of planned lunar missions. The ESA's Terrae Novae 2030+ program~\cite{TerraeNovae}, aligned with the exploration strategy Explore2040~\cite{EXPLORE2040}, outlines a detailed plan for future lunar exploration. They emphasize the importance of crater detection as a key component in achieving the goal of sending the first European astronaut to the lunar surface. ESA's lunar lander program, Argonaut~\cite{Argonaut}, will enable Europe to access the Moon as a fully European project. It consists of two main components: the Lunar Descent Element (LDE) and the Passenger. The LDE is responsible for both transporting and landing the Passenger on the lunar surface.  Since even the smallest craters, only a few meters in diameter, can potentially cause a lander to tip over, their accurate detection conditions the success of the whole exploration mission. The main focus in recent crater detection research has been on detecting large craters~\cite{zou2024}. In this work, we overcome this limitation by targeting craters of various sizes and shapes, also under varying sun incidence angles, resulting in craters that range from barely to highly shadowed. 

According to Chaini et al.~\cite{Chaini2024} crater detection algorithms (CDAs) can be divided into manual methods, where craters are labeled by humans, and in automatic crater detection approaches. Manual crater detection has the limitation of being very time consuming and error-prone~\cite{tewari2023}. Further, according to Robbins et al.~\cite{robbins2014}, manual crater count can vary by up to 40\%. Hence, the focus of CDAs is on automatic crater detection, where traditional methods such as edge detection~\cite{Emami2015} have been used. In recent years, deep learning has become the primary focus in the field of CDAs, as the data provided by NASA's Lunar Reconnaissance Orbiter Camera (LROC) contains complex information, including varying crater shapes and sizes, diverse lighting conditions, and rugged terrain, where traditional approaches face significant limitations~\cite{Chaini2024}. Several deep learning approaches have been proposed, e.g. Tewari et al.~\cite{tewari2022}, that used a Mask R-CNN model with a ResNet-50 backbone and a feature pyramid network, trained on optical, elevation, and slope map inputs for crater detection. In a later work, Tewari et al.~\cite{tewari2024} proposed a novel crater shape retrieval system that first estimates crater boundaries using an unsupervised adaptive rim extraction algorithm on DEM data. Subsequently, they refine these boundaries through cascaded Mask R-CNNs in a semi-supervised manner, enabling highly accurate crater shape retrieval. Several researchers incorporated various versions of the You Only Look Once (YOLO) model as their CDA. Nan et al.~\cite{nan2025} proposed YOLOv8-LCNET, an enhanced YOLOv8-based model for automatic lunar crater detection, which integrates a Partial Self-Attention (PSA) mechanism in the backbone and a Gather-and-Distribute (GD) module in the neck to improve multi-level feature information and detection accuracy. La Grassa et al.~\cite{lagrassa2023}, Mu et al.~\cite{Mu2023} and Zhu et al.~\cite{Zhu2023} also employed variants of the YOLO model and achieved good crater detection results. Zhang et al.~\cite{zhang2024} focused on small-scale crater detection using an anchor-free deep learning approach. Specifically, they applied CenterNet with transfer learning, detecting crater centers as peaks in a heat map and directly regressing their sizes,  without relying on non-maximum suppression (NMS). Jia et al.~\cite{Jia2023} introduced the AE-TransUNet+, an improved hybrid Transformer network based on the TransUNet~\cite{transunet}, which achieved high accuracy in detecting small craters. Other researches, such as Silburt et al.~\cite{Silburt2019}, Mao et al.~\cite{Mao2022} and Lee et al.~\cite{Lee2021} adapted the U-Net~\cite{Ronneberger2015} architecture for the task of crater detection. With CraterDANET~\cite{Yang2022} the authors introduced  a framework based on domain adaptation, whereby they trained their model solely on auto-annotated synthetic crater data before transferring it to real-world images. This approach achieved competitive results without the need for labelled real-world data. The authors of~\cite{Liu2024fasterrcnn} demonstrated that improving the Faster R-CNN model~\cite{Ren2015} is very effective for detecting craters larger than 200 m in diameter on the Kaguya TC morning map, which has a resolution of 7.403 m/pixel. Jia et al.~\cite{Jia2021} proposed SCNeSt, a split-attention network with self-calibrated convolutions, that can be applied for multi-source data for detecting craters on the Moon, Mercury and Mars. A recent publication introduced a variant of the DEtection TRansformer (DETR)~\cite{detr} with Crater-DETR~\cite{Crater-DETR}, that is specialized for small crater detection yielding good results on a dataset consisting of Mars craters. Leveraging pretrained foundation models and adapting them to specific downstream tasks has also proven promising~\cite{Bommasani2022}. For example, Giannakis et al.~\cite{giannakis2024} applied SAM~\cite{SAM} for crater detection and achieved promising results, even without fine-tuning it on remote sensing images containing craters. For further existing CDA research, we refer to the review article of Chaini et al.~\cite{Chaini2024}. 

In recent years, Vision Transformers (ViTs) have played a vital role in the realm of computer vision. ViTs are based on the Transformer~\cite{Vaswani2017} architecture, treating images as sequences of tokens in a similar way to how Transformers treat sentences. Tasks such as object detection, semantic segmentation, and image classification are increasingly addressed using ViTs~\cite{Dong2025}. The patch-based processing structure of ViTs enables them to capture long-range dependencies across an image, enabling them for tasks that demand a comprehensive understanding of complex visual content~\cite{Khan2022}. Through positional encoding and multi-head self-attention (MSA), ViTs have demonstrated strong performance on various vision benchmarks, often matching or outperforming convolutional neural networks (CNNs) in tasks such as object detection~\cite{Shah2023}. However, unlike CNNs, ViTs lack strong inductive biases and thus typically require extensive pre-training on large-scale datasets to generalize effectively. Extensive pre-training on large datasets enhances model performance and, as noted by Dosovitskiy et al.~\cite{Dosovitskiy2020}, can even surpass the benefits of inductive biases. This flexibility and scalability make ViTs a robust and versatile foundation for advanced methods, including feature extraction and task-specific adaptation.  While both the Transformer achieved remarkable results in the field of Natural Language Processing (NLP)~\cite{Xu2023} and the ViT in the realm of computer vision~\cite{Saha2025}, their combination paved the path for Vision-Language Models (VLMs)~\cite{Tao2025} with the introduction of CLIP~\cite{radford2021}. CLIP is designed to align paired text and images within a shared embedding space. Since then, many vision-language based models have been proposed in recent years, such as the OWL-ViT~\cite{minderer2022} (Vision Transformer for Open-World Localization) and its successor, OWLv2~\cite{minderer2024}. 
According to~\cite{Li2024}, VLMs offer new opportunities in the field of remote sensing, such as scene classification, image captioning, visual question answering and object detection. The recent trend suggests growth in VLM adaptations for remote sensing. For instance, Liu et al.~\cite{Liu2024} introduced RemoteCLIP, a model that can be applied to various downstream tasks such as zero-shot image classification, image-text retrieval and object detection. RemoteCLIP serves as the first vision-language foundation model in the realm of remote sensing, as noted by Tao et al.~\cite{Tao2025}. Zhang et al.~\cite{Zhang2023RS5MAG} proposed a framework for adapting VLMs to remote sensing. This involved introducing a domain-specific vision-language model and a large-scale RS5M image-text dataset. Using this framework, they fine-tuned CLIP to create GeoRSCLIP, achieving significant improvements in downstream tasks such as zero-shot classification, cross-modal text-image retrieval and semantic localization. As VLMs are becoming more widely adopted in the remote sensing domain~\cite{Tao2025}, but are not yet adapted for the specific task of crater detection to the best of our knowledge, our aim is to contribute to this area by adapting a VLM-based detector for crater detection in remote sensing imagery. Additionally, as many researchers employ CNN-based models for crater detection~\cite{Chaini2024}, according to Zhang et al.~\cite{Zhang2022}, ViT-based models generalize better than CNNs when faced with a distribution shift. Since the available lunar surface data contains a broad range of illumination conditions and various terrain conditions, we opt for a ViT-based model. Current limitations in the field of crater detection are as follows: Zhang et al.~\cite{zhang2024} provided a promising crater detection model, but emphasized that they need to further increase the generalization capabilities of their model in a future work. Furthermore, the detection of small lunar craters remains challenging. As Fan et al.~\cite{Fan2022} outlined in their work, even though they provided an efficient crater detection algorithm, they still face limitations in detecting small craters.  Zou et al.~\cite{zou2024} also emphasized the scarcity of small lunar crater detection. 
 
Previous work~\cite{Bauer2025} demonstrated a promising few-shot crater detection approach using the OWLv2 model by customizing the similarity metric utilizing the high-dimensional image embeddings.  In this work, we build on that approach by fine-tuning the OWLv2 model for the task of crater detection on lunar surface images. As stated by Bommasani et al.~\cite{Bommasani2022}, pre-training foundation models on large-scale datasets and subsequently fine-tuning them for specific downstream tasks is a widely adopted paradigm in computer vision. To address the computational and memory challenges of full fine-tuning, several parameter-efficient fine-tuning (PEFT) techniques have been proposed in the literature, including Adapters~\cite{houlsby2019parameter}, Prefix-Tuning~\cite{li2021prefix}, Prompt Tuning~\cite{lester2021power}, BitFit~\cite{zaken2022bitfit}, AdapterFusion~\cite{pfeiffer2021adapterfusion}, and IA$^3$~\cite{liu2022few}. These approaches enable efficient task-specific adaptation while keeping most of the base model parameters fixed. For our methodology, we adopt LoRA~\cite{LoRA}, that inserts low-rank trainable matrices into the MSA layers of frozen ViT weights and in the detection heads, allowing efficient fine-tuning by optimizing only a small number of additional parameters.

\subsection*{Contributions}

Detecting craters on the lunar surface is an important research objective, and the ESA's ambitions for lunar exploration demand a reliable CDA that performs well across diverse lunar regions and imaging conditions. Building on the contributions in~\cite{Bauer2025}, which employed a few-shot crater detection strategy using the OWLv2 model and introduced a penalty term in the similarity score calculation, we propose a deep-learning based enhancement, summarized as follows:
\begin{itemize}
    \item[(i)] We propose a method that fine-tunes the pre-trained OWLv2 model by introducing additional trainable parameters with LoRA, enabling efficient adaptation to the crater detection task. The model is trained by minimizing a combined loss of Complete Intersection over Union (CIoU) for localization and contrastive loss for classification.
    \item[(ii)]  The vision-language model is trained and evaluated on a manually labeled dataset from the IMPACT project~\cite{Impact}. We observe strong performance in detecting craters of varying sizes and shapes, supported by both qualitative and quantitative evaluation. We demonstrate, that the model has strong generalization capabilities as well as the ability to detect small lunar craters.
    \item[(iii)] We compare our fine-tuned model with other state of the art (SoTA) models. Specifically, we train three variants of the YOLO~\cite{YOLO2016} series as well as RT-DETR~\cite{Wenyu2023} and compare it with our proposed method. Further we show that fine-tuning also improves the quality compared to our previous work in~\cite{Bauer2025} and compared to the zero-shot capabilities of the pre-trained OWLv2 model.
\end{itemize}

\section{METHODOLOGY}~\label{methodology}

In this section, we present our method for fine-tuning the OWLv2~\cite{minderer2024} model on a manually annotated dataset from the IMPACT~\cite{Impact} project for a novel accurate crater detection approach.  
\subsection{Vision Transformer in Computer Vision}~\label{attention}
With the introduction of Transformers~\cite{Vaswani2017} and the incorporation of attention modules, a new paradigm in NLP emerged. Attention-based models enable the modeling of dependencies between any elements in the input or output sequence, regardless of their distance~\cite{Vaswani2017}.
In this process, words or subwords are tokenized and embedded into a high-dimensional vector space. Positional encodings are added to the input embeddings to provide the model with information about the order of tokens~\cite{Vaswani2017}. Each layer applies multi-head self-attention by projecting the input tokens with weight matrices $W_Q, W_K, W_V$ onto the queries $Q$, keys $K$, and values $V$. Attention is further the scaled dot-product
\begin{equation}
    \text{Attention}(Q,K,V) = \text{softmax}(\frac{QK^T}{\sqrt{d_k}})V,
\end{equation}
computed across multiple heads, where $\frac{1}{\sqrt{d_k}}$ denotes the scaling factor. The head outputs are then concatenated and projected with $W_O$. A position-wise feed-forward network follows, with residual connections and layer normalization around each sublayer, yielding the Transformer output tokens~\cite{Vaswani2017}. The original Transformer is structured as an encoder-decoder architecture.
Several benchmark models such as BERT~\cite{Devlin2019} and GPT-3~\cite{Brown2020} have been developed, with GPT-3 trained on 45 TB of text data and containing approximately 175 billion parameters. Pre-trained models that are Transformer-based are capable of achieving state-of-the-art performances on various tasks~\cite{Lin2022}. As a consequence, the Transformer has become the standard architecture in NLP. Following the remarkable success of the Transformer in NLP, researchers began adapting the architecture to computer vision tasks such as image classification, object detection, and segmentation.
Its first major breakthrough in image classification was achieved by the Vision Transformer (ViT)~\cite{Dosovitskiy2020}, which applies an encoder-only Transformer architecture directly to sequences of image patches. The idea is that an image is divided into fixed-size non-overlapping patches, and each patch, similarly to words or subwords in a standard Transformer, is treated as a token and embedded as a high-dimensional vector, with positional encodings added to retain spatial information~\cite{Dosovitskiy2020}. In general, a class token is prepended to all tokens and serves as an overall representation of the image. Similar to large pre-trained Transformer models in NLP, several foundation models based on ViTs have emerged in computer vision, including CLIP~\cite{radford2021} (Contrastive Language-Image Pretraining), SAM~\cite{SAM} (Segment Anything Model), and DINO~\cite{DINO} (Self-Distillation with No Labels). The development of VLMs was paved by the combination of large, pre-trained Transformer-based models in NLP and ViT-based models in computer vision. CLIP is a pioneer in advancing VLMs by bridging the gap between the two fields~\cite{Tao2025}. Designed to align text and visual information in a shared embedding space, CLIP is capable of tasks such as zero-shot image classification through extensive contrastive pre-training on large amounts of image-text data~\cite{radford2021}. This means it is not limited to a fixed set of labels or categories seen during pre-training. Building on these developments, OWL-ViT~\cite{minderer2022} extends CLIP to the setting of object detection.

\subsection{OWLv2: Vision Transformer for Open-World Localization}

OWL-ViT~\cite{minderer2022} builds upon the CLIP~\cite{radford2021} architecture by combining a ViT~\cite{Dosovitskiy2020} for image encoding and a Transformer-based text encoder. These components are jointly pre-trained on 3.6 billion image-text pairs using contrastive learning to align visual and textual features in a shared embedding space. For object detection, OWL-ViT removes the token pooling layer, attaches classification and box regression heads to the Transformer output tokens, and fine-tunes the model on roughly 2 million object-level annotated images. OWLv2~\cite{minderer2024} extends OWL-ViT by incorporating large-scale self-training: The open-vocabulary object detector OWL-ViT CLIP-L/14~\cite{minderer2022} is used to generate pseudo-boxes over WebLI~\cite{webli} ($\sim$ $10$ billion image-text pairs), and models are self-trained on a $\sim$ $2$ billion image subset. A new objectness head improves efficiency by selecting tokens that are most likely to correspond to actual objects against the background, reducing unnecessary computations during detection.

OWLv2 processes the input images by first resizing them to a fixed image size of $960\times 960$ pixels with bilinear interpolation. The image is then divided into 3600 non-overlapping patches of size $16\times 16$ pixels. Each patch is linearly projected into a 768-dimensional vector space, and positional encodings are added to preserve spatial structure. A special class token that captures global image information, is prepended to the sequence, resulting in a total of 3601 tokens. These tokens are then passed through the ViT encoder, which models local and global semantic relationships via MSA, analogous to token processing in NLP transformers.
After encoding, the resulting embeddings capture local and global semantic relationships and are used for object detection. The class token plays a key role in aggregating global context. As noted by Minderer et al.~\cite{minderer2022}, multiplying the class token with the patch embeddings can enhance object detection performance. Each of the 3600 patch embeddings is passed through three distinct prediction heads. The box head generates a four-dimensional vector representing the predicted bounding box coordinates. The objectness head produces a scalar score indicating the likelihood that an object is present in the corresponding bounding box or indicating background. The class head outputs a 512-dimensional image class embedding that encodes the visual features of the predicted object.

Simultaneously, the text encoder processes objects of interest provided as textual prompts (e.g., ``dog'', ``cat'', etc.). The output are 512-dimensional vectors representing the corresponding objects of interest. The core idea behind contrastive ViTs is that objects described by text should have a similar representation, based on cosine similarity, to the 512-dimensional image class embeddings derived from the visual features. Models like OWL-ViT and OWLv2 are capable of performing both zero-shot and few-shot object detection~\cite{minderer2022,minderer2024}.
Zero-shot object detection refers to the model's ability to detect and localize objects from classes that were not necessarily seen during training. Instead of relying on visual examples for each class, the model utilizes semantic information provided through text prompts or class descriptions~\cite{Tan2021}. Few-shot object detection, on the other hand, addresses the challenge of recognizing novel classes using only a limited number of labeled examples. Hence, OWLv2 provides a framework for adapting zero- and few-shot object detection to the task of crater detection.

\subsection{OWLv2 for Crater Detection}
\begin{figure}[t]
  \centering
  \begin{tabular}{cccc}
      \includegraphics[width=0.13\textwidth]{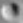} &
      \includegraphics[width=0.13\textwidth]{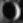} &
      \includegraphics[width=0.13\textwidth]{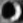} &
      \includegraphics[width=0.13\textwidth]{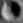} \\
  \end{tabular}
  \caption{Examples of craters for the few-shot crater detection approach.}\label{fig: groundtruthcatalog}
\end{figure}

The naive approach of detecting craters using the zero-shot object detection approach using the pretrained OWLv2 model and prompting the word ``crater'' presents several challenges. First of all, it is not clear what the exact prompt should be, as alternatives like ``craters'', ``small craters'', ``impact craters'', or even ``circles'' are all plausible, since craters appear circular when viewed from a nadir perspective~\cite{cadogan2020}. Moreover, applying this approach revealed further limitations as the confidence scores indicating whether detected objects are actual craters were consistently low. This may be explained by the fact that the OWLv2 model was not specifically trained on remote sensing images of the Moon. While the word ``crater'' may occur in captions, it was likely not aligned with lunar craters as visual objects, which limits OWLv2's ability to detect them reliably.  In prior work~\cite{Bauer2025}, a few-shot approach was found more suitable by passing example craters (cf. Fig.~\ref{fig: groundtruthcatalog}) and their corresponding image class embeddings, then identifying similar objects, i.e. craters,  based on cosine similarity. In this setup, only the ViT path of the OWLv2 model was used, while the text Transformer was not considered. The analysis of embeddings classified as craters revealed high similarity even between visually distinct instances, such as brightly illuminated craters and those almost entirely shadowed. This indicates a high intraclass similarity among the embeddings describing craters. However, the presence of similarly high interclass similarity, i.e., between craters and non-craters, motivated the introduction of a penalty mechanism to improve separation in the embedding space. Further, as noted by~\cite{Weng2025}, the performance of frozen ViTs pretrained on natural images is often limited when directly applied to remote sensing tasks due to a significant domain gap, making further adaptation or fine-tuning typically necessary. Similarly, Luo et al.~\cite{Luo2024}, in their adaptation of the Segment Anything Model (SAM)~\cite{SAM} to remote sensing, emphasized this domain gap and underscored the importance of domain-specific adaptation strategies.  Thus, fine-tuning the OWLv2 model on lunar-specific remote sensing data is crucial to fully exploit its capability for detecting  lunar craters. 

\begin{figure*}[!ht]
  \centering
  \includegraphics[width=\textwidth]{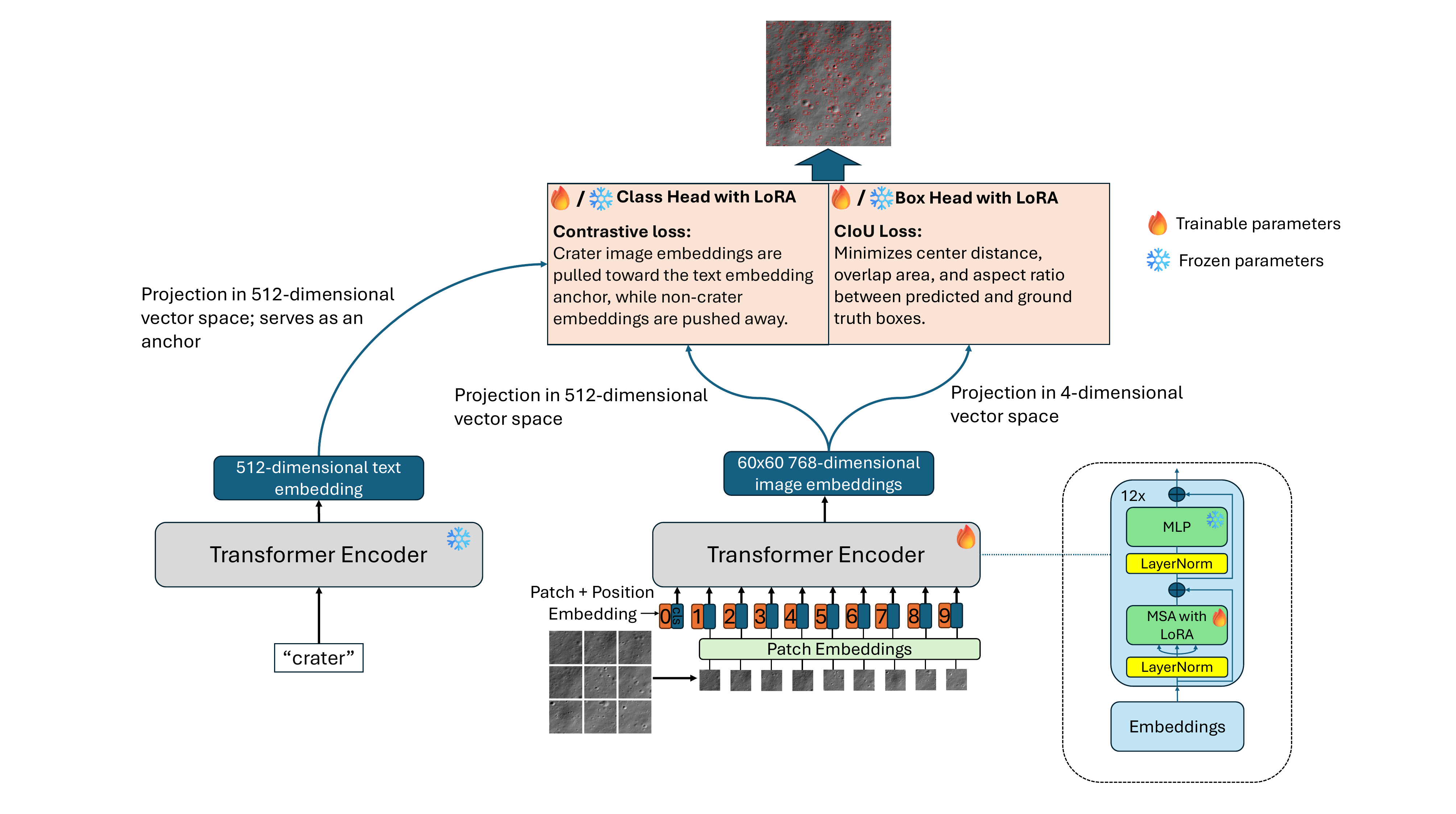}
  \caption{Schematic overview of the proposed crater detection method: Calibrated Data Record (CDR) images are processed by the OWLv2 Vision Transformer (ViT), where trainable LoRA parameters are inserted into the encoder. Simultaneously, the word ``crater'' is encoded by the frozen text Transformer and used as an anchor in the shared embedding space. The Box Head minimizes a CIoU-based loss between predicted boxes and ground truth, while the Class Head applies a contrastive loss to separate crater from non-crater embeddings utilizing the anchor vector, by adding LoRA parameters into both heads for efficient fine-tuning.  We remove the objectenss head of the original OWLv2 model, because it is of no need for our specific task of crater detection. For an ablation study, we keep the box head and the class head in trained and untrained states, respectively.}\label{fig: scheme}
\end{figure*}
 
In this approach, we utilize both the text Transformer and the ViT as shown in Fig.~\ref{fig: scheme}. The $512$-dimensional embedding for the word ``crater'' serves as an anchor in a contrastive framework: crater image embeddings are pulled toward this anchor, while non-crater embeddings are pushed away, by maximizing the cosine similarity for craters and minimizing it for non-craters. The OWLv2 model processes $3600$ patches per image, allowing detection of up to $3600$ objects. 

To assign predictions to ground truth (GT) craters, we use the Hungarian matching algorithm~\cite{Kuhn1955}, which minimizes the  cost function based on CIoU~\cite{ciou} (complete intersection over union). Let $y = \{y_i\}_{i=1}^M$ denote the set of GT craters for a given image, where $M$ is the number of annotated craters, and let $\hat{y} = \{\hat{y}_j\}_{j=1}^N$ be the set of $N$ model predictions, with $M \ll N$ (in OWLv2, $N=3600$; median $M_m$ of annotated craters in our GT dataset: $M_m = 200$). Following~\cite{detr}, we pad $y$ with $(N-M)$ ``no-object'' entries so that both sets have equal cardinality and we denote the predictions by $i$ (rows) and the padded GT boxes by $j$ (columns). We compute a pairwise CIoU cost matrix $C\in\mathbb{R}^{N\times N}$,
\begin{equation}
    C_{ij} = 1 - \mathrm{CIoU}(\hat{b}_i, b_j),
\end{equation}
where $\hat{b}_i \in [0,1]^4$ and $b_j\in [0,1]^4$ are predicted and GT boxes, respectively.  
We then obtain the alignment
\begin{equation}
    \sigma^* = \operatornamewithlimits{argmin}_{\sigma \in S_N}
 \sum_{i=1}^{N} C_{i,\sigma(i)},
\end{equation}
where $S_N$ is the set of all permutations of $N$ elements.  
The problem is solved efficiently with the Hungarian algorithm~\cite{Kuhn1955}. After matching, we compute the bounding box loss only for the $M$ true GT and prediction pairs:
\begin{equation}
    \mathcal{L}_{\text{box}}=\frac{1}{M}\sum_{i=1}^{M}\bigl[1-\mathrm{CIoU}\,(\hat{b}_{\sigma^{*}(i)}, b_{i})\bigr],
\end{equation}
where $\mathcal{L}_{\text{box}}$ is the same CIoU-based regression loss used to define the matching cost.

Further, let $t\in\mathbb{R}^{512}$ denote the $\ell_2$-normalized text embedding of the word ``crater'', and let $e_j\in\mathbb{R}^{512}$ denote the $\ell_2$-normalized image class embedding for prediction $j$.
For each image, we mark the $\sigma^{*}(i)$ indices found by the Hungarian assignment of the matching result as positives and all other indices as negatives.
We then apply a cosine-similarity contrastive loss that
(i) pulls the positive image embeddings $e_{\sigma^{*}(i)}$ toward the anchor $t$ and
(ii) pushes the remaining embeddings toward orthogonality:
\begin{align}
\mathcal{L}_{\text{cls}}
&=
\frac{1}{P}\sum_{j\in\mathcal{P}} \bigl(1-\cos(e_j,t)\bigr)
\notag  \\
&+\frac{1}{N}\sum_{j\in\mathcal{N}} \max\!\bigl(0,\,\cos(e_j,t)-\tau\bigr),
\end{align}
with $y_j\in\{0,1\}$ the binary label derived from $\sigma^{*}$,
$\mathcal{P}=\{j \mid y_j=1\}$ and $\mathcal{N}=\{j \mid y_j=0\}$ denote the
sets of positive and negative pairs with $P=|\mathcal{P}|$ and
$N=|\mathcal{N}|$, respectively, and $\tau$ is a margin for the negatives
(here, $\tau=0.1$). That means, we let the model learn what a (visual) crater embedding $e_j\in\mathbb{R}^{512}$ should look like given the vector $t\in\mathbb{R}^{512}$ that corresponds to the textual description ``crater''. 
In total, we minimize the loss function 

\begin{equation}
    \mathcal{L}_{\text{total}} = \alpha_\text{box} \mathcal{L}_\text{box} + \alpha_\text{cls} \mathcal{L}_\text{cls},
\end{equation}
where we set $\alpha_\text{box} = 7.5$ and $\alpha_\text{cls} = 3.0$ to prioritize precise localization while still ensuring strong classification performance; these values were chosen empirically. After the training process, the crater detection pipeline follows the zero-shot object detection principle. By prompting the word ``crater'' through the text Transformer and passing out-of-distribution (OOD) selected CDR images through the ViT, the model predicts only objects with 512-dimensional vectors similar to the vector representing ``crater''.

\subsection{Injecting parameters with LoRA}

As described in Subsection~\ref{methodology}~\ref{attention}, the queries, keys, and values are computed by linearly projecting the input embeddings: $Q = W_Q x, K = W_K x, V = W_V x$. During fine-tuning, these weight matrices are frozen. Mathematically, for $m \in \{Q, K, V\}$, a pre-trained weight matrix $W_m \in \mathbb{R}^{d\times k}$ is updated by adding trainable low-rank matrices $A_m \in \mathbb{R}^{r\times k}, B_m \in \mathbb{R}^{d\times r}$, $r \ll \min(d,k)$. For an input $x$ a corresponding step of the forward path yields
\begin{equation}
    W_m x + \Delta W_m x = W_m x + B_m A_m x.
\end{equation}
Note that during training, the pre-trained weight matrix $W_m$ remains frozen and is not updated. In the image encoder, we apply LoRA weight matrices only to $W_Q$ and $W_V$, as following the procedure in~\cite{LoRA}, and use a rank of $r = 8$. Traditionally, LoRA parameters are injected only into the multi-head self-attention (MSA) and feed-forward (MLP) layers of the Transformer. However, since our objective is to learn and represent crater-specific features within the high-dimensional embedding space, we also inject LoRA parameters into the class and box head instead of fully fine-tuning them, following the idea in~\cite{yuan2025}, where LoRA parameters were added in the classification head. To enable this, we apply LoRA to both heads with various ranks of $r \in \{4, 8, 16\}$ to conduct an ablation study. One study only fine-tuned the ViT, while keeping the heads frozen. For the downstream task of crater detection with the OWLv2 model, we emphasize that fine-tuning the objectness head is unnecessary and would only add computational and memory overhead to the training process. Therefore, we remove the objectness head and focus solely on fine-tuning the box and classification heads. In doing so, only a small fraction of the OWLv2 model parameters are trainable.

\section{EXPERIMENTAL SETUP}
\begin{figure*}[t]
    \centering
    \begin{minipage}[t]{0.25\textwidth}
        \centering
        \includegraphics[width=\linewidth]{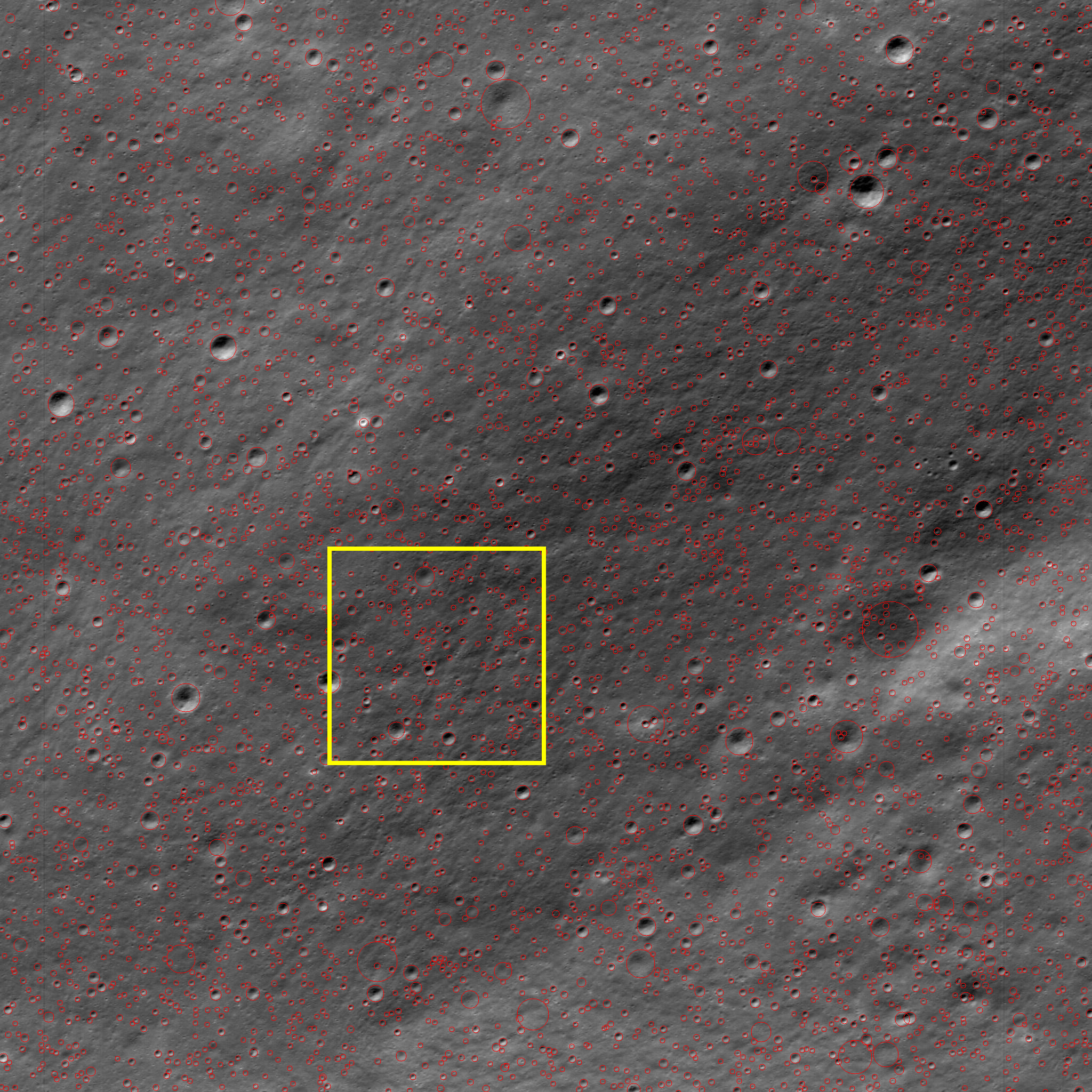}
        \par\vspace{4mm}
        \includegraphics[width=0.5\linewidth]{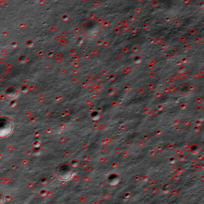}
    \end{minipage}
    \hspace{3mm}
    \begin{minipage}[t]{0.25\textwidth}
        \centering
        \includegraphics[width=\linewidth]{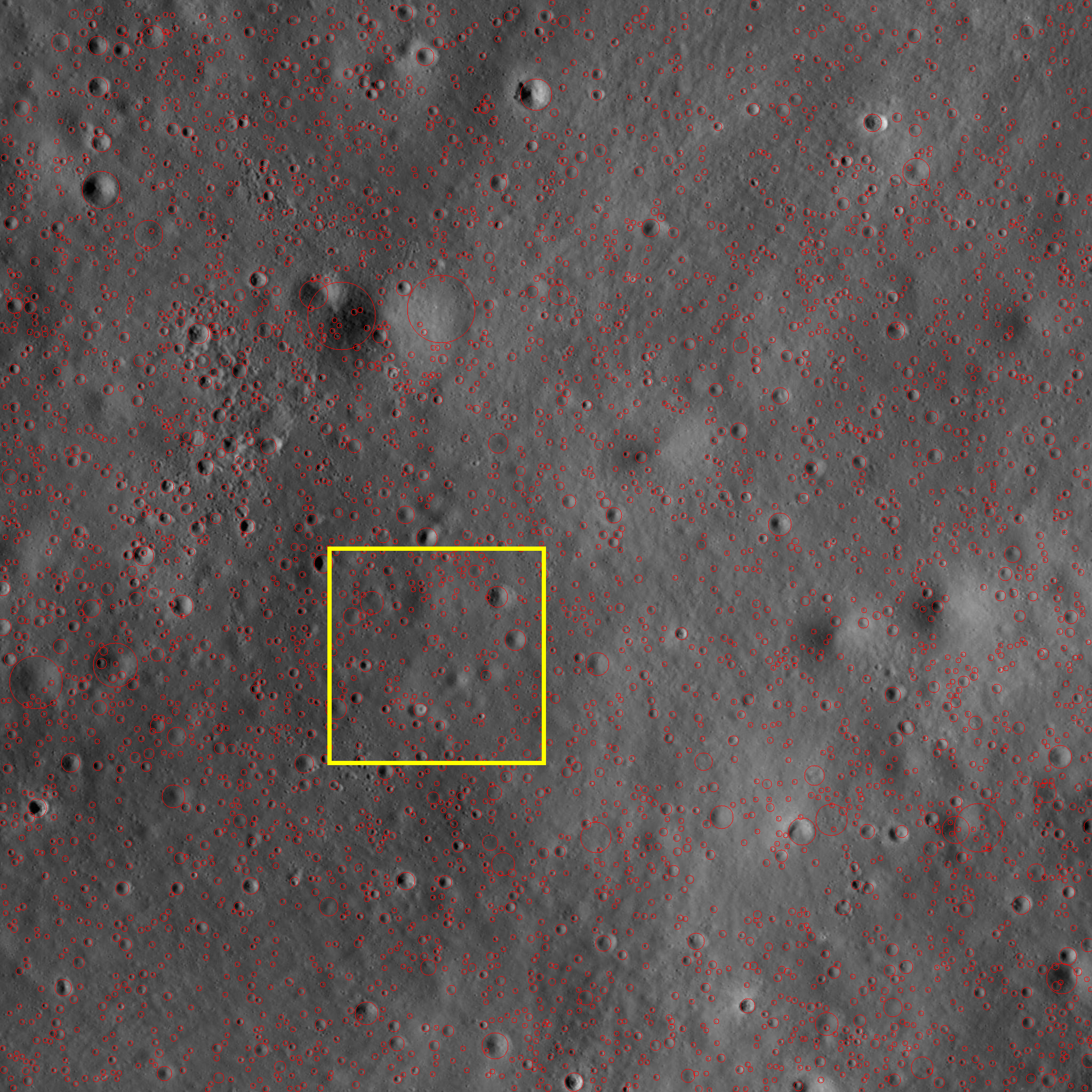}
        \par\vspace{4mm}
        \includegraphics[width=0.5\linewidth]{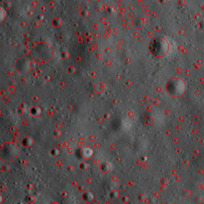}
    \end{minipage}
    \hspace{3mm}
    \begin{minipage}[t]{0.25\textwidth}
        \centering
        \includegraphics[width=\linewidth]{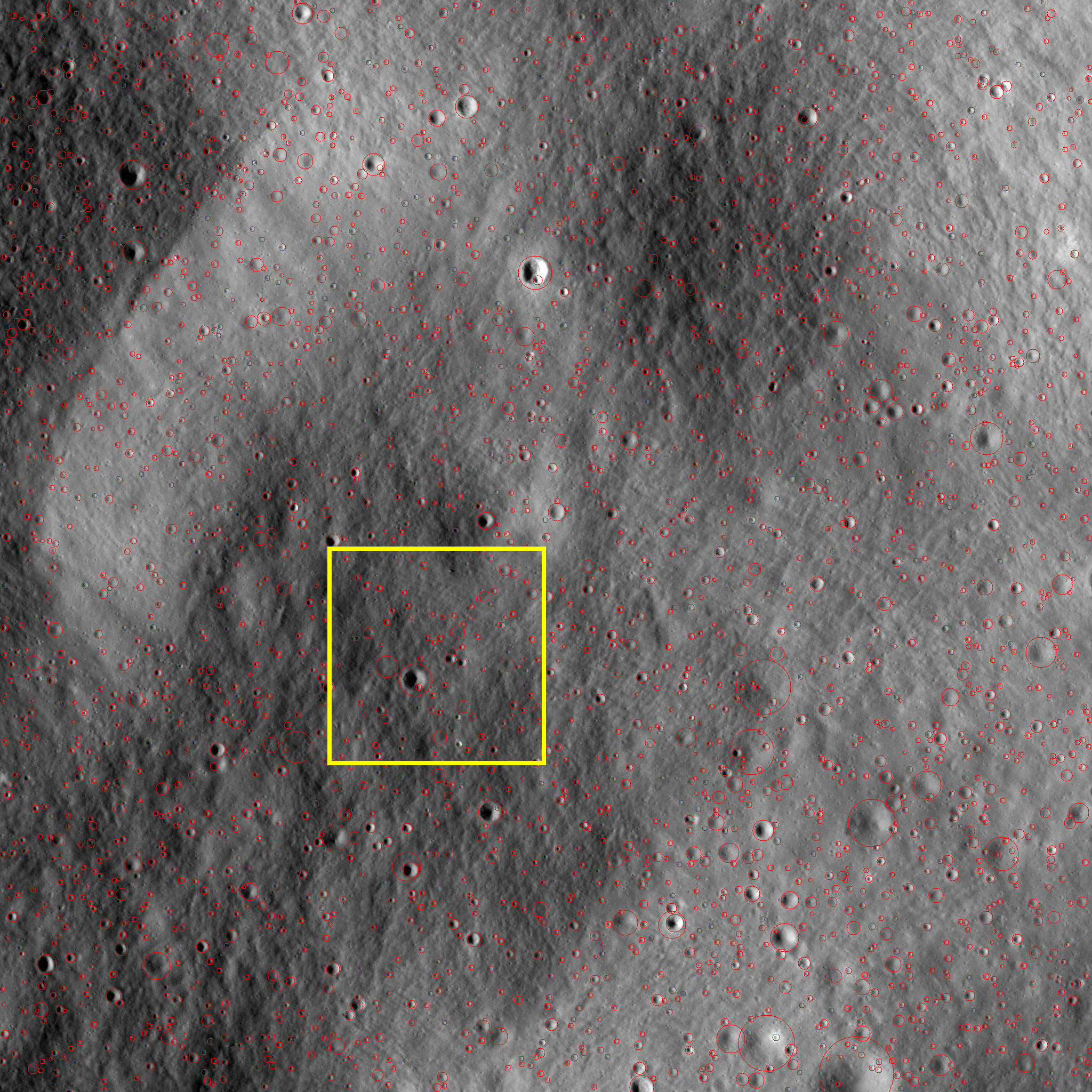}
        \par\vspace{4mm}
        \includegraphics[width=0.5\linewidth]{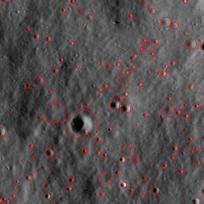}
    \end{minipage}
    \caption{
        Example images of the underlying IMPACT data containing crater annotations on calibrated data record images from the Lunar Reconnaissance Orbiter Camera. These images were used to train and evaluate our proposed method. Each column: (top) Original $2048\times 2048$ image with a yellow rectangle indicating the zoomed region.  (bottom) Cropped and zoomed-in region corresponding to the yellow rectangle above. The red circles indicate the annotated ground truth craters. For training and evaluation, we tile each of the IMPACT images onto 16 tiles of size 512$\times$512. 
    }
    \label{fig:annotations_good_quality}
\end{figure*}
\subsection{Dataset}
NASA launched the Lunar Reconnaissance Orbiter (LRO) in June 2009 and since then it has been observing the Moon from an altitude of 50 km, providing various images from the Moon and its surface~\cite{Robinson2010}. To do so, it has several instruments onboard such as the Lunar Reconnaissance Orbiter Camera (LROC). The LROC consists of two distinct camera systems, the Narrow Angle Cameras (NACs) and the Wide Angle Camera (WAC). The two NACs, the NAC-Left (NAC-L) and NAC-R (NAC-Right), were designed to capture high-resolution images down to $\sim$ 0.5 m/pixel, and the WAC provides images at a scale of $\sim$ 100 m/pixel~\cite{Robinson2010}. The mission's objectives include identifying potential landing sites, detecting surface hazards, and creating high-resolution maps of the lunar polar regions. For the task of detecting craters, we select images from the LROC Calibrated Data Record (CDR), that have undergone radiometric and geometric corrections, and downloaded them from the Planetary Data System archive (\url{https://pds.lroc.asu.edu/data/}). 

For fine-tuning the OWLv2 model, we use the manually labeled dataset provided by the IMPACT project~\cite{Impact}. The project provides a game, where players are encouraged to annotate lunar craters as accurately as possible while building a base on the lunar surface.  As a result, the project publishers, who plan to make the dataset publicly available in the near future (currently only available to ESA), have collected and verified a large number of crater annotations spanning various lunar regions. Example annotations are shown in Fig.~\ref{fig:annotations_good_quality}. Crater annotations are provided as normalized circle coordinates ($c_x, c_y, r$) relative to the CDR image size of $2048 \times 2048$ pixels, with craters excluded that are smaller than 8 pixels in diameter. Each image is identified by its CDR image ID (e.g., M118673590LC), followed by the $(x_{\text{min}}, y_{\text{min}})$ coordinates of the upper left corner of image tile. The final number indicates the tile size, which is always 2048 (for example: M118673590LC\_1508\_36964\_2048). For training and evaluation, we convert these circular coordinates to square bounding boxes, cut each image in 16 tiles of size $512\times 512$  and use them as GT. This  also motivates the use of LoRA in the box head: fully unfreezing the box head would lead the model to associate craters strictly with square shapes, which could hinder generalization to other celestial bodies or to  specific emission angles, where craters can appear more elliptical and require rectangular shaped bounding boxes.

\subsection{Processed IMPACT Dataset} 

\begin{figure*}[t]
  \centering
  \subfloat[\label{fig:bad_annotations_a}]{
    \begin{minipage}[t]{0.30\textwidth}\centering
      \includegraphics[width=\linewidth]{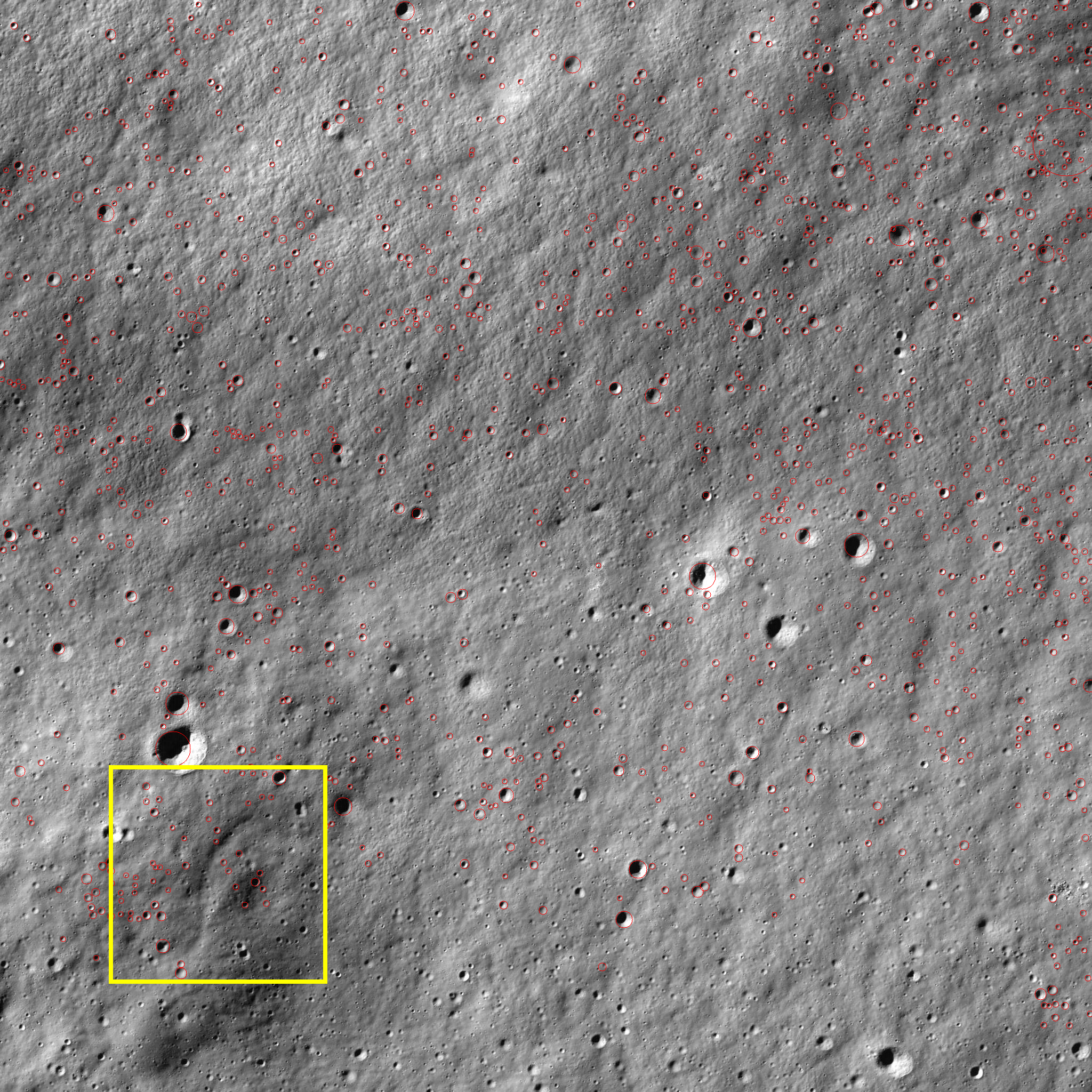}\\[4mm]
      \includegraphics[width=0.33\linewidth]{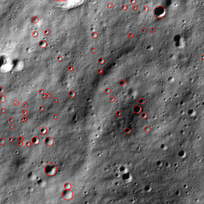}
    \end{minipage}
  }\hspace{3mm}
  \subfloat[\label{fig:bad_annotations_b}]{
    \begin{minipage}[t]{0.30\textwidth}\centering
      \includegraphics[width=\linewidth]{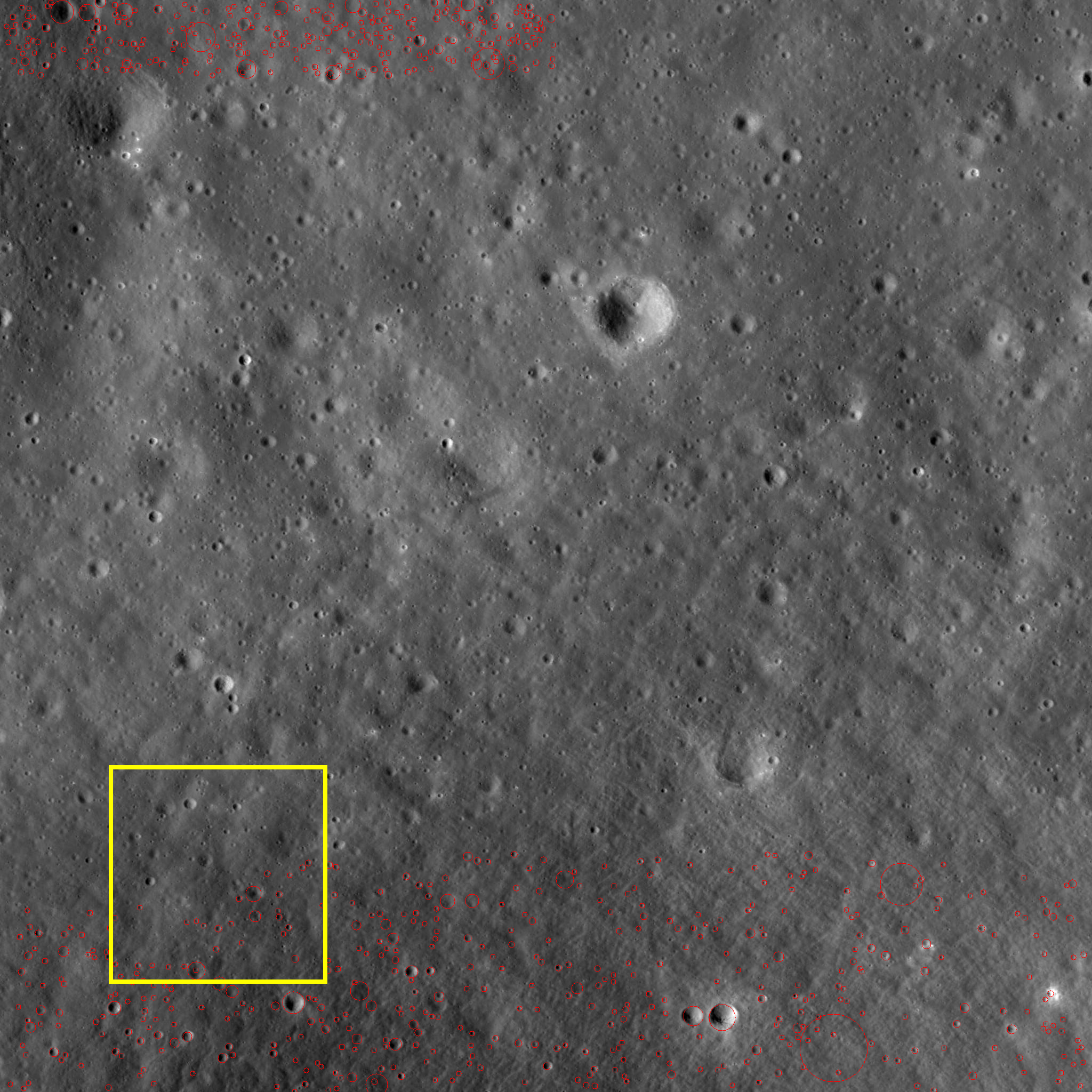}\\[4mm]
      \includegraphics[width=0.33\linewidth]{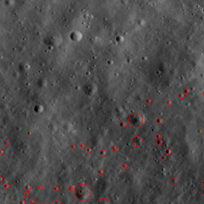}
    \end{minipage}
  }
  \caption{The IMPACT dataset is a manually annotated lunar crater database required by a large number of IMPACT participants. This database is used for training and evaluation and requires preprocessing and preselection, as indicated in this Figure. We show examples of poorly annotated images from the annotated IMPACT data used as ground truth. Each column: top: original $2048\times 2048$ image with a yellow rectangle marking the zoomed region; bottom: cropped zoom corresponding to that rectangle. Red circles show annotated ground-truth craters. Images of that poor quality are removed and not used for training and evaluation.}\label{fig:poorly_annotated_images}
\end{figure*}

Since players are encouraged to label craters and, as previously noted, manual crater counts can vary widely, the developers introduced a metric to quantify annotation accuracy. Each crater receives an accuracy score based on the consistency of markings from multiple players. Regions incorrectly labeled as craters, where only a few players mark the same location, receive low accuracy scores. Hence, the raw data contains many false positive annotations, making data pre-processing essential to obtain a reliable dataset for training and validation. Mostly following the developers' recommendations,  we set an accuracy threshold of 0.55, slightly below the actually suggested value of 0.6, and remove noisy annotations that cover more than 85\% black pixels to enhance dataset quality. This step is necessary because regions in full shadow frequently confused users, leading them to mark image noise. We consider any pixel with an intensity lower than a grayscale value of 30 (in the range [0, 255]) as black. Further visual inspection indicates that some images contain inconsistent labeling, with many craters missing. Therefore, we removed entire images that were of insufficient quality, cf. Fig.~\ref{fig:poorly_annotated_images}. As noted by Robbins et al.~\cite{Robbins2019}, the decision to label a feature as a crater depends on how conservative or liberal the annotator is. Since our goal is to support ESA's lunar landing efforts, we prioritize the recall metric, aiming to minimize false negatives, and exclude images with missing annotations. Consequently, we adopt a more liberal approach to labeling features as craters. After preprocessing, we retain a total of 880 tiles of size $512 \times 512$, containing a total of 178,812 crater annotations. We divide the dataset based on the following criteria. Since in  some cases, multiple tiles are collected from the same image ID but from different regions, for instance, two distinct images may correspond to a single image ID (e.g. M118673590LC\_1508\_36964\_2048 and M118673590LC\_1508\_41060\_2048), we ensure that all images from the same image ID are placed in the same dataset split — train, validation, or test — to prevent data leakage. We target an $80 - 10 - 10$ split for training, validation, and testing, respectively. We group all $2048 \times 2048$ tiles by their parent LROC image ID and randomly assign approximately 80\% of image IDs to the training set, and approximately 10\% each to validation and testing, ensuring that no CDR image ID is shared across splits. At the image level, the split results in $42 / 7 / 6$ images for train, validation, and test, corresponding to approximately 76\%, 13\%, and 11\%. The actual tile counts deviate slightly from the intended $80 - 10 - 10$ split. We consider this modest drift acceptable given the strict grouping by image ID.

\subsection{Data Augmentation and Implementation Details}
\begin{table*}[t]
\centering
\caption{Data augmentation sub-policies and their parameters. Op: Operation, 
P: probability, RoM: Range of Magnitude. 
Following~\cite{Zoph2020}, each sub-policy consists of two operations with an associated probability and magnitude. 
We keep the continuous ranges provided by \texttt{Albumentations}~\cite{Albumentations}, which directly reflect our implementation.}
\setlength{\tabcolsep}{4pt}
\begin{tabular}{lcccccc}
\hline
& \multicolumn{3}{c}{\textbf{Operation 1}} & \multicolumn{3}{c}{\textbf{Operation 2}} \\
\cline{2-4} \cline{5-7}
\textbf{Sub-policy} & \textbf{Op} & \textbf{P} & \textbf{RoM} & \textbf{Op} & \textbf{P} & \textbf{RoM} \\
\hline
Sub-policy 1 & TranslateX & 0.6 & [-0.4, 0.4] & Equalize & 0.8 & — \\
Sub-policy 2 & BBox\_Only\_Rotate & 0.2 & [-180, 180] & Scale & 1.0 & [-0.3, 0.3] \\
Sub-policy 3 & ShearY & 0.6 & [-10, 10] & BBox\_Only\_Rotate & 0.6 & [-180, 180] \\
Sub-policy 4 & Rotate & 0.6 & [-30, 30] & ColorJitter & 1.0 & br.=0.6, co.=0.5, sa.=0.5, hu.=0.1 \\
Sub-policy 5 & No operation & — & — & No operation & — & — \\
\hline
\end{tabular}\label{dataaugmentation}
\end{table*}
To increase the size of the training set and to get a more diverse dataset we apply data augmentation strategies. We follow the idea presented in~\cite{Zoph2020} by constructing five different augmentation policies. Each policy is selected randomly and within each policy, individual augmentation operations are applied with a fixed probability and a certain range of magnitude. We keep the exact same data augmentation policies as in~\cite{Zoph2020}, but replace the \texttt{BBox\_Only\_TranslateY} with  a \texttt{BBox\_Only\_Rotate} technique to randomize the shadow orientation and discourage the model from associating a specific shadow direction with the crater class. In addition, we replace \texttt{Cutout} with \texttt{Scale} to better capture craters at various sizes. An overview over the data augmentation strategy is provided in Table~\ref{dataaugmentation}. For each image in the train dataset we apply one of the five policies and double the training set size. Training is performed for 100 epochs with a batch size of 4. We apply a weight decay of $1 \times 10^{-3}$ and use the AdamW~\cite{adamW} optimizer. The learning rate is linearly warmed up from 0 to $1 \times 10^{-4}$ during the first 10\% of epochs, then decayed to zero over the remaining epochs following a cosine schedule. The experiments were conducted using an NVIDIA H100 NVL GPU with 94 GB of VRAM.

\section{RESULTS}

In this section, we evaluate our approach both qualitatively and quantitatively. 
The quantitative evaluation is based on recall and precision, defined as
\begin{align}
    \text{Recall} &= \frac{\text{TP}}{\text{TP + FN}},\\
    \text{Precision} &= \frac{\text{TP}}{\text{TP + FP}},
\end{align}
where TP denotes the number of true positives (actual craters), FN the number of false negatives (missed craters), and FP the number of false positives (incorrectly identified as craters). We consider a detection to be a true positive (TP) if the IoU is greater than 0.40. This IoU threshold is increased by 0.1 compared to that chosen by Emami et al.~\cite{Emami2019}, Yang et al.~\cite{Yang2022} and Tewari et al.~\cite{tewari2022} for crater detection. In inference, we apply non-maximum suppression with a IoU threshold of 0.12 to remove duplicate boxes for the same crater.

\subsection{Quantitative Evaluation on IMPACT}
\begin{table*}[t]
\centering
\caption{Ablation study on LoRA placement and rank in different components of the model. The average metrics are computed on the IMPACT test set. The IoU threshold for true positives is set to 0.4.}\label{tab:ablation_heads}
\setlength{\tabcolsep}{4pt}
\begin{tabular}{l|ccc|ccc|cc}
\hline
\multirow{2}{*}{Method}
& \multicolumn{3}{c|}{Setting}
& \multicolumn{3}{c|}{LoRA Rank}
& \multicolumn{2}{c}{Performance} \\

& ViT Encoder & Classification Head & Box Head 
& $r_{\text{cls}}$ & $r_{\text{box}}$ & $r_{\text{ViT}}$ 
& avg. Recall (\%) & avg. Precision (\%) \\

\hline

LoRA & \checkmark & —          & —          & —  & —  & 8 & 86.2 & 51.3 \\
LoRA & \checkmark & \checkmark & \checkmark & 4  & 4  & 8 & 85.5 & 50.4 \\
LoRA & \checkmark & \checkmark & \checkmark & 8  & 8  & 8 & 85.4 & 51.1 \\
LoRA & \checkmark & \checkmark & \checkmark & 16 & 16 & 8 & 85.1 & 50.0 \\

\hline
\end{tabular}
\end{table*}

First, we conduct an ablation study of the corresponding LoRA ranks within the model on the test dataset. For the ViT, the class head and the box head, we consider the respective ranks, $r_\text{ViT}, r_\text{cls}, r_\text{box}$. We fix $r_\text{ViT} = 8$, while considering varying ranks for $r_\text{cls}, r_\text{box} \in \{4, 8, 16\}$. We also consider a case in which the respective heads are not fine-tuned and are kept frozen. Table~\ref{tab:ablation_heads} shows that fine-tuning only the ViT, while keeping the heads frozen, yields slightly better results. This may indicate that additional training of the prediction heads is not required, whereas the ViT backbone must be adapted to address the domain gap, while the heads are already capable to classify and localize the craters. Nevertheless, the results are all rather similar. Henceforth, to discuss the results, we continue with the model, where  the heads are kept frozen.
The quantitative evaluation is performed on the test dataset of 6 images and therefore 96 tiles of size $512 \times 512$ in total. To calculate recall and precision, we convert the GT labels, which are given as circular coordinates, into squares enclosing the circular annotations. As the pretrained model is trained on rectangular bounding boxes, we keep the detections as rectangles and compare them with the GT circles that have been transformed into squares for the metric calculation. In Table~\ref{tab:recall_precision} we outline the corresponding evaluation results. The recall values range from $80.3\%$ to $92.6\%$ with an average of $86.2\%$, while the precision values range from $28.0\%$ to $71.4\%$ with an average of $51.3\%$. While the recall on each test image is relatively high, indicating a good detection rate of true positives, the precision is in comparison rather low. We emphasize that many detections not present in the GT, counted by definition as false positives, are in fact plausible crater detections.

\begin{table}[t]
\caption{Recall and precision of the fine-tuned OWLv2 model on the test dataset with $r_\text{ViT} = \{8\}$, and the box and class head kept frozen.}
\centering
\begin{tabular}{p{0.62\textwidth}cc}
\toprule
\textbf{Image Name} & \textbf{Recall (\%)} & \textbf{Precision (\%)} \\
\midrule
M1184106708LC\_1508\_34916\_2048 & 84.9 & 51.0 \\
M1184106708LC\_1508\_47204\_2048 & 82.8 & 28.0 \\
M1250049562LC\_1508\_16484\_2048 & 90.6 & 45.5 \\
M1466265041LC\_1508\_100\_2048   & 92.6 & 49.5 \\
M1466265041LC\_1508\_12388\_2048 & 80.3 & 71.4 \\
M1466265041LC\_1508\_49252\_2048 & 85.7& 62.4 \\
\bottomrule
\end{tabular}
\label{tab:recall_precision}
\end{table}

\begin{figure*}[!ht]
    \centering
    \subfloat[M1184106708LC\_1508\_34916\_2048]{
        \includegraphics[width=0.27\textwidth]{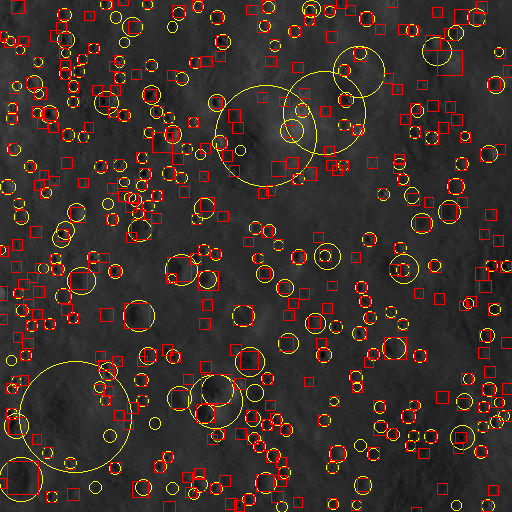}
    }\hspace{3mm}
    \subfloat[M1184106708LC\_1508\_47204\_2048]{
        \includegraphics[width=0.27\textwidth]{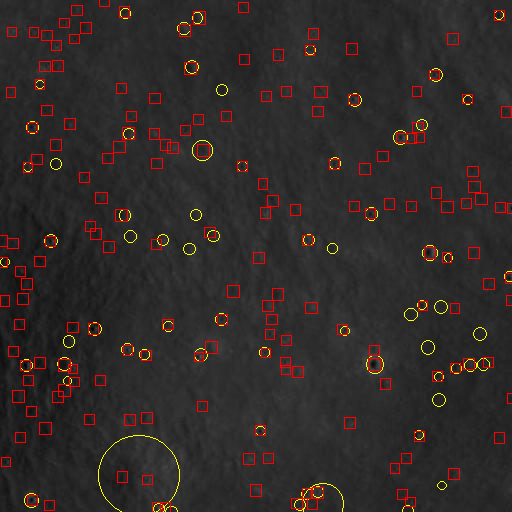}
    }\hspace{3mm}
    \subfloat[M1250049562LC\_1508\_16484\_2048]{
        \includegraphics[width=0.27\textwidth]{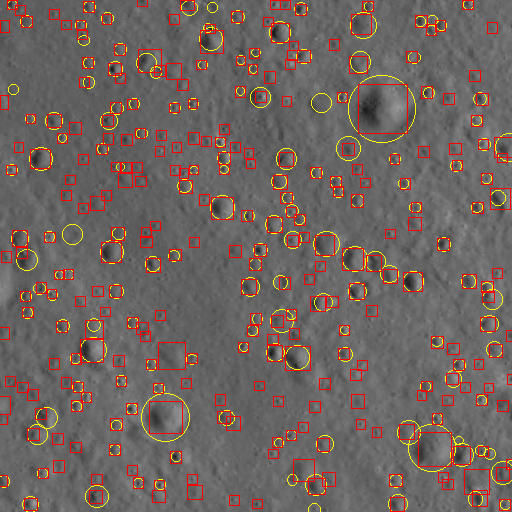}
    }\\[3mm]
    \subfloat[M1466265041LC\_1508\_100\_2048]{
        \includegraphics[width=0.27\textwidth]{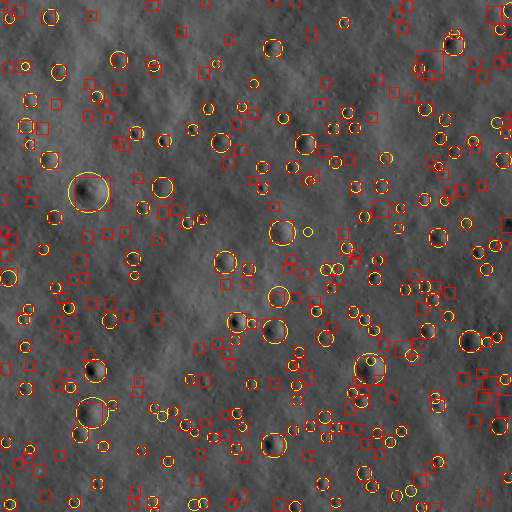}
    }\hspace{3mm}
    \subfloat[M1466265041LC\_1508\_12388\_2048]{
        \includegraphics[width=0.27\textwidth]{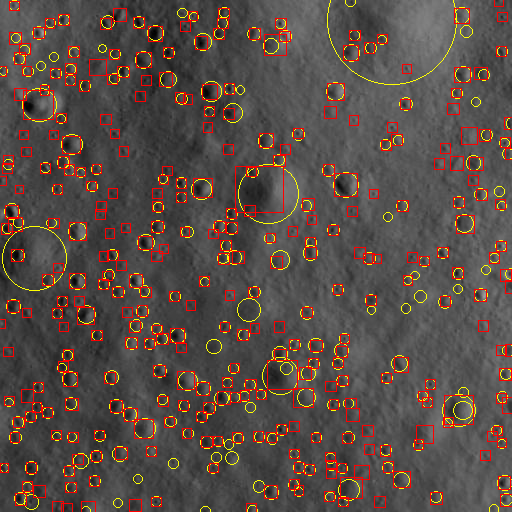}
    }\hspace{3mm}
    \subfloat[M1466265041LC\_1508\_49252\_2048]{
        \includegraphics[width=0.27\textwidth]{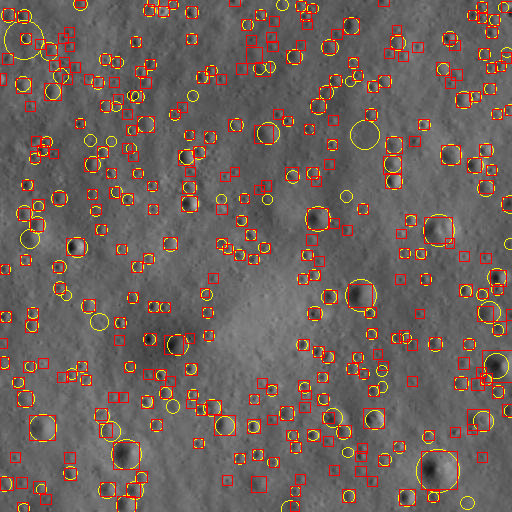}
    }
    \caption{Qualitative results on one $512 \times 512$ tile from each of the six images from the IMPACT test dataset. Yellow circles indicate ground truth annotations, and red bounding boxes denote model predictions. The model predicts more craters than are annotated in the ground truth. We emphasize, although some false positives occur, most of these additional detections are actual craters that were not annotated.}\label{fig:qualitative_evaluation_test_dataset}
\end{figure*}

This is also highlighted in Fig.~\ref{fig:qualitative_evaluation_test_dataset}, where we evaluated our method on the test set with GT annotations. We observe a high consistency between our predictions (red boxes) and the GT annotations (yellow circles). Notably, in almost every case, the number of predictions exceeds the number of GT annotations. This can also be explained by the \texttt{scale} data augmentation technique, where a negative scale makes the GT craters smaller during training and causes the model to predict craters smaller than 8 pixels. This effect is also evident in Fig.~\ref{fig:qualitative_evaluation_test_dataset}, where most of the newly detected craters appear to be small. This is not a limitation of our method, since our goal is to detect craters across all sizes and not to exclude small ones. In our work, we aimed to construct a more liberal catalog out of the IMPACT annotations,  to reduce the number of missed detections. Nevertheless, many craters were still absent from the GT annotations, partly due to the applied size constraint of 8 pixels in diameter and the accuracy threshold.  As a result many detections counted as false positives are, in fact, likely true craters that were simply not labeled,  a challenge also noted by Tewari et al.~\cite{tewari2024} and Silburt et al.~\cite{Silburt2019}.

\subsection{Qualitative Evaluation}
\begin{figure*}[!ht]
    \centering
    \includegraphics[width=0.24\textwidth]{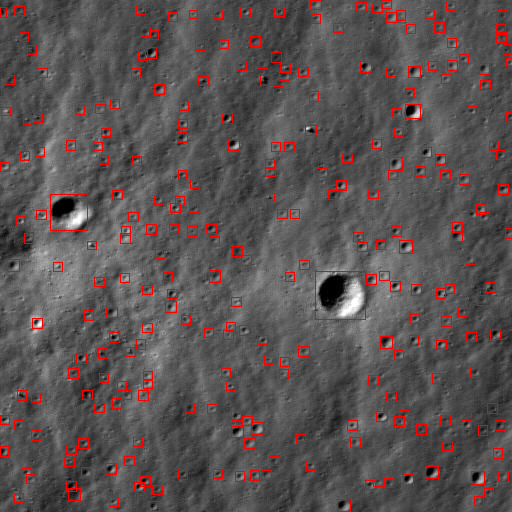}\hspace{1mm}
    \includegraphics[width=0.24\textwidth]{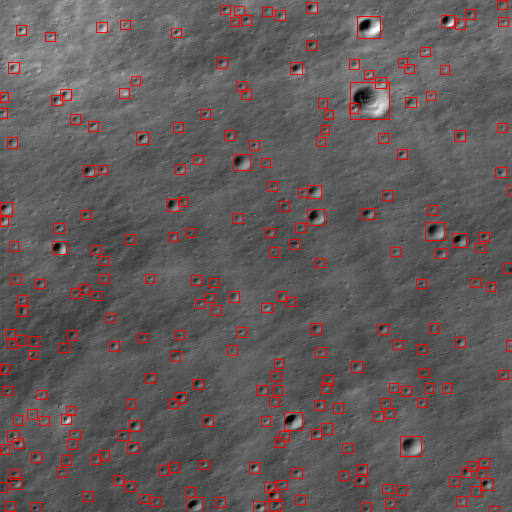}\hspace{1mm}
    \includegraphics[width=0.24\textwidth]{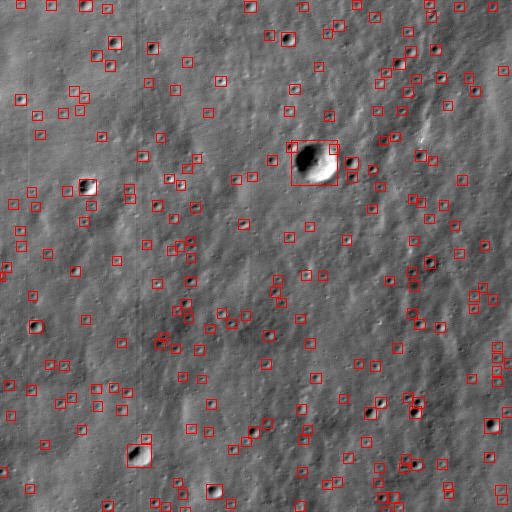}
    \\[1mm]
    \includegraphics[width=0.24\textwidth]{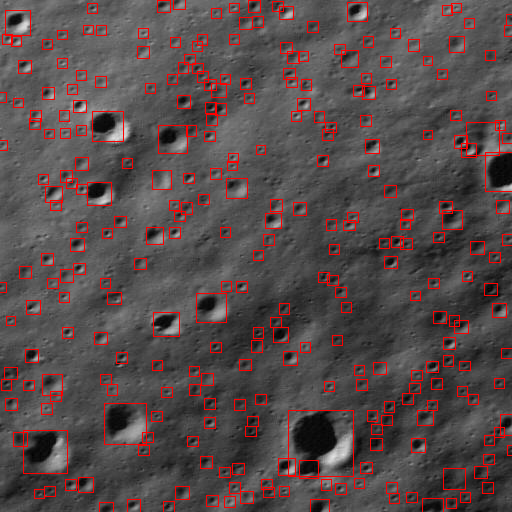}\hspace{1mm}
    \includegraphics[width=0.24\textwidth]{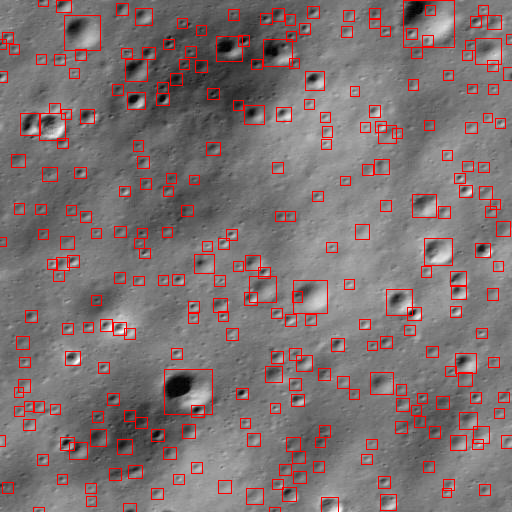}\hspace{1mm}
    \includegraphics[width=0.24\textwidth]{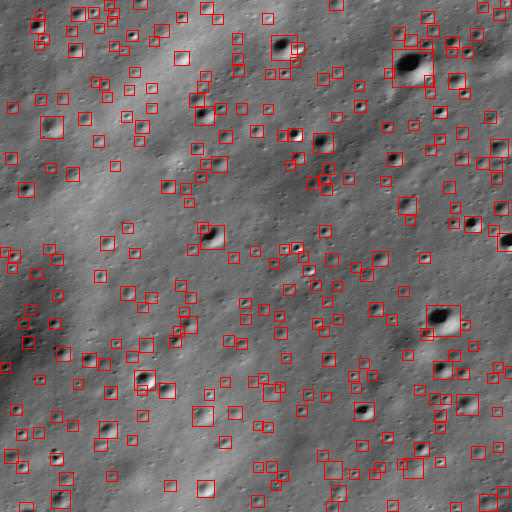}
    \\[1mm]
    \includegraphics[width=0.24\textwidth]{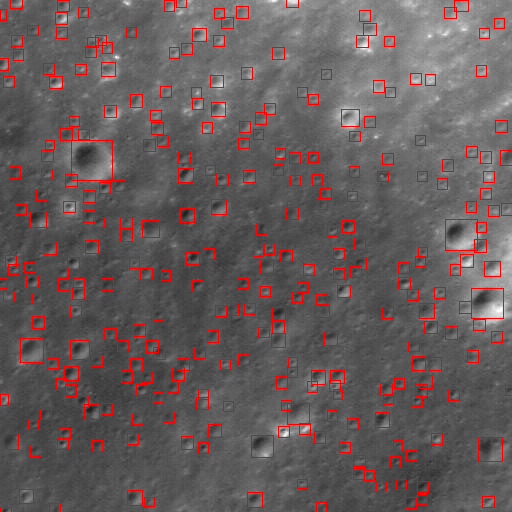}\hspace{1mm}
    \includegraphics[width=0.24\textwidth]{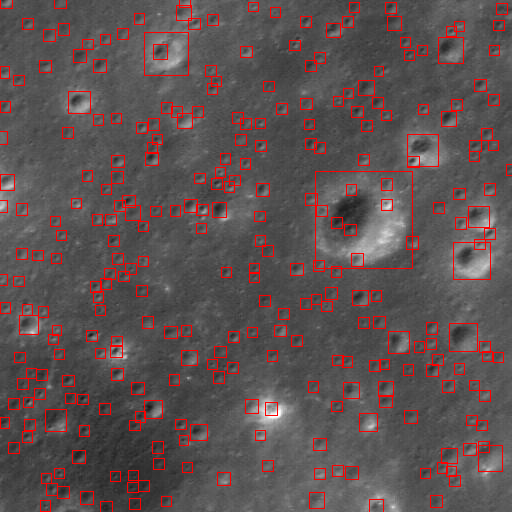}\hspace{1mm}
    \includegraphics[width=0.24\textwidth]{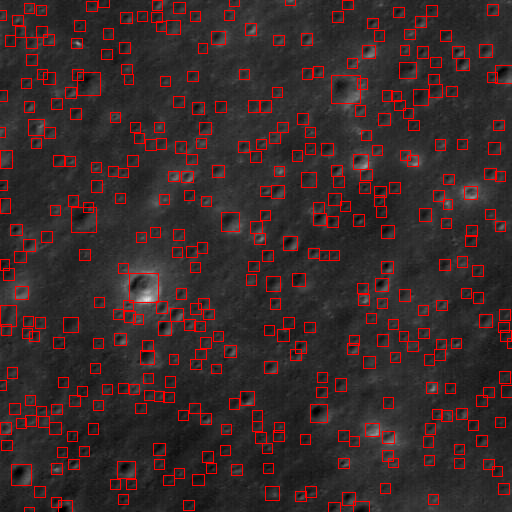}
    \\[1mm]
    \includegraphics[width=0.24\textwidth]{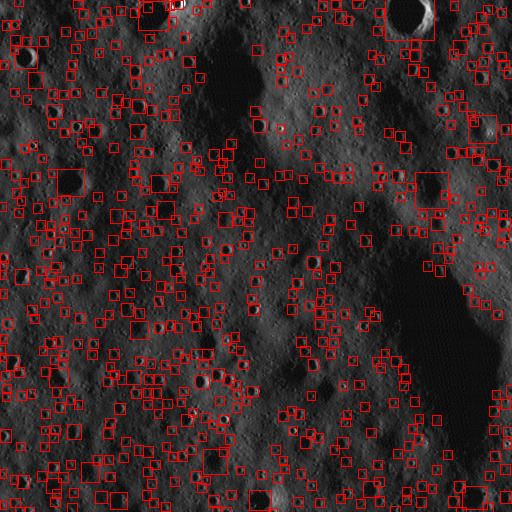}\hspace{1mm}
    \includegraphics[width=0.24\textwidth]{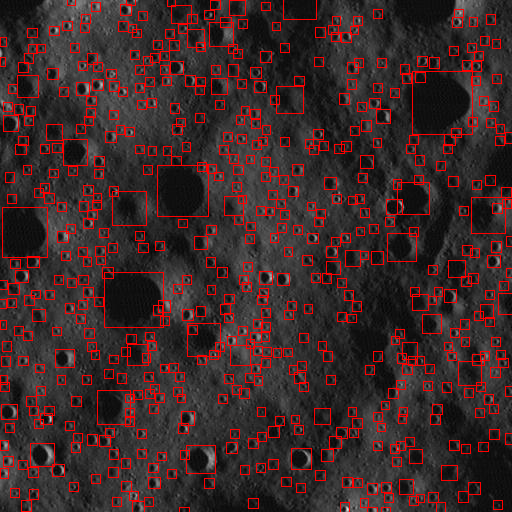}\hspace{1mm}
    \includegraphics[width=0.24\textwidth]{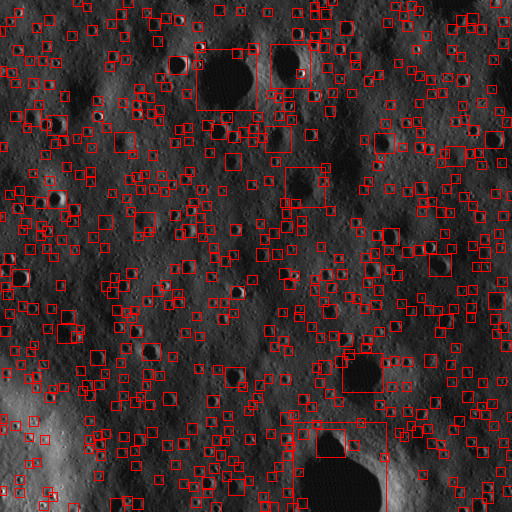}

    \caption{Qualitative results on CDR images unseen during both training and validation, selected randomly. Each row consists of three tiles of size $512 \times 512$. First row: M1369716293R, second row: M1310840621L, third row: M1501897277L, fourth row: M1496678308R. Despite varying image conditions, such as incidence angles ranging from 40.2 to 85.1 degrees, an overall sufficient detection quality is achieved. In the fourth row, especially for larger craters, the bounding boxes often capture only the shadowed part of the crater, that also appear circular. Under these illumination conditions, however, even for human observers the crater contours are difficult to extract.}\label{fig:qualitative_evaluation_unseen_cdr}
\end{figure*}

\begin{figure*}[!ht]
    \centering
    \includegraphics[width=0.3\textwidth]{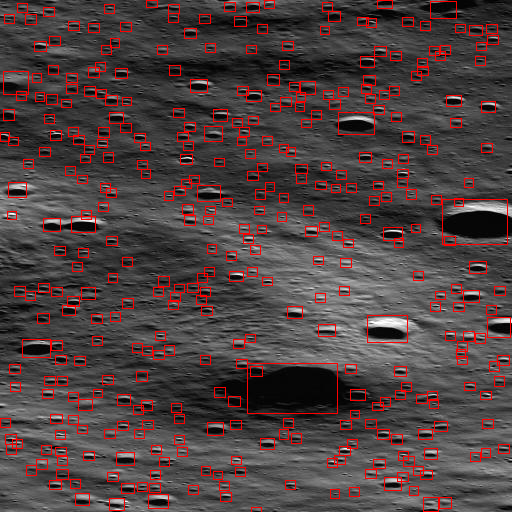}\hspace{2mm}
    \includegraphics[width=0.3\textwidth]{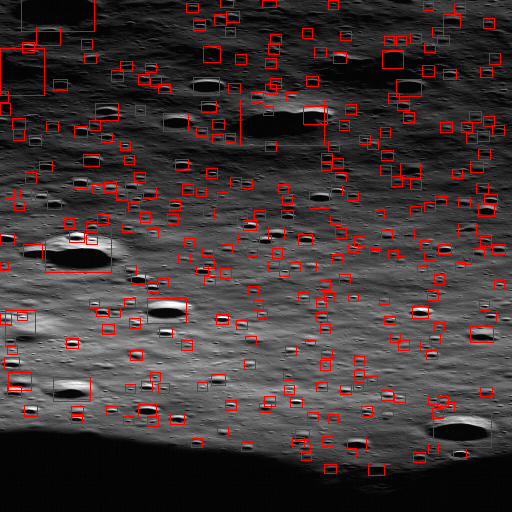}\hspace{2mm}
    \includegraphics[width=0.3\textwidth]{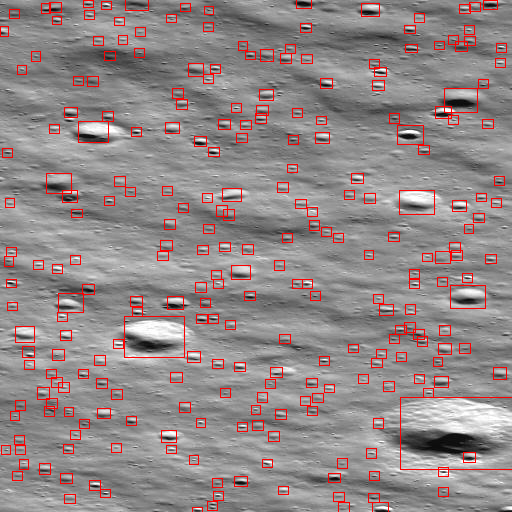}
    \caption{Qualitative results on CDR image M185603505L from the lunar south pole, unseen during both training and validation. Due to the emission angle, the craters appear more elliptical than circular yielding more rectangular shaped bounding boxes, a condition completely absent from training and validation. Nevertheless, during inference the craters are detected sufficiently well, highlighting the strong generalization capability of the fine-tuned OWLv2 model.}\label{fig:qualitative_evaluation_unseen_cdr_southpole}
\end{figure*}

In Fig.~\ref{fig:qualitative_evaluation_unseen_cdr}, we observe good visual results in detecting craters across various lunar regions, that were not seen during training or validation. These images were selected as out-of-distribution (OOD) samples from the CDR archive and the experiments were conducted on tiles of size $512 \times 512$. For a robust evaluation, we selected images with different resolutions, ranging from 0.49 m/px to 1.85 m/px, as well as varying illumination incidence angles ranging from 40.2 to 85.1 degrees, to demonstrate the model's applicability across diverse imaging conditions. Our results show that the method detects craters of various sizes and shapes and remains robust under varying image conditions. Considering the detections in Fig.~\ref{fig:qualitative_evaluation_unseen_cdr_southpole}, which showcases an image of the lunar south pole, where craters can appear more elliptical, our approach is not limited, as the craters are still detected and represented with more rectangular bounding boxes. However, we emphasize that it is not possible to detect the locations and shapes of craters precisely without further fine-tuning on off-nadir lunar data. During training, the model has only seen squared-shapded labelled data from a nadir view. We demonstrate the model's generalization capabilities by applying it to off-nadir images and showing that it can detect craters that appear differently in these images despite being elliptical. However, the model's box head does not accurately predict their shape, as it should be more rectangular-shaped and enclose the full crater. Further, we emphasize that our method does not produce many visually observable false positives. For nearly every detected crater, one could argue that it represents an actual crater, even if no corresponding GT annotation exists. 
Overall we note that the qualitative evaluation of our method yields that it is of sufficient quality. 
Nevertheless, we highlight the need for further methodological improvements, as in some cases, only the shadowed region of a crater is detected. This effect is even more evident in merely illuminated regions, where the shadowed areas appear circular, as shown in the last row of Fig.~\ref{fig:qualitative_evaluation_unseen_cdr}. Under such illumination conditions, it is difficult even for human observers to extract the exact crater boundaries.

\subsection{Comparison with Other Methods}

In order to compare our approach to other SoTA object detection models, we utilize several versions of the YOLO series~\cite{YOLO2016} and fully fine-tune them on the same IMPACT data as we fine-tune the OWLv2 model. Specifically, we train YOLOv8n, YOLOv11m and YOLOv26m~\cite{ultralytics_yolo}. Additionally, we train RT-DETR~\cite{Wenyu2023}, a real-time end-to-end object detection model based on DETR~\cite{detr}. To demonstrate the crater detection capabilities of OWLv2 resulting from the proposed fine-tuning strategy, we compare its crater detection capabilities quantitatively on the test set of IMPACT and on one OOD example image with those of the pre-trained OWLv2 zero-shot approach, the OWLv2 few-shot approach introduced in~\cite{Bauer2025}, the YOLO models and the RT-DETR.

\begin{figure*}[!ht]
    \centering
     \subfloat[original tile]{\includegraphics[width=0.3\textwidth]{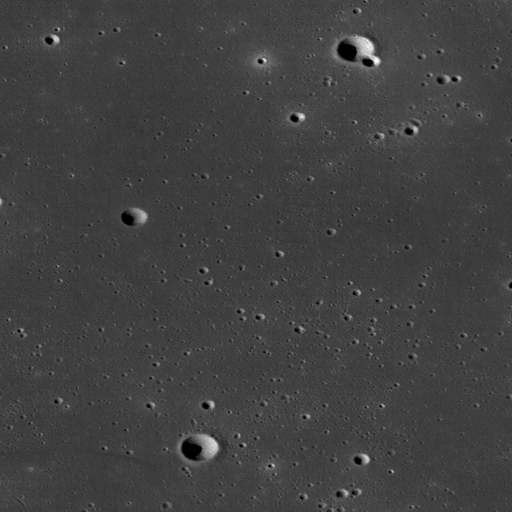}}\hspace{2mm}
    \subfloat[OWLv2 zero-shot]{\includegraphics[width=0.3\textwidth]{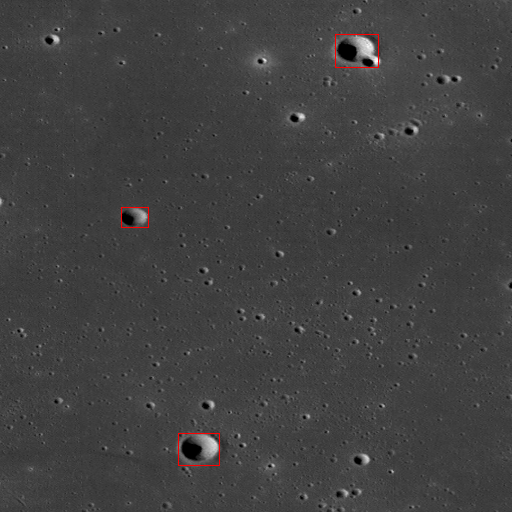}}\hspace{2mm}
    \subfloat[OWLv2 few-shot]{\includegraphics[width=0.3\textwidth]{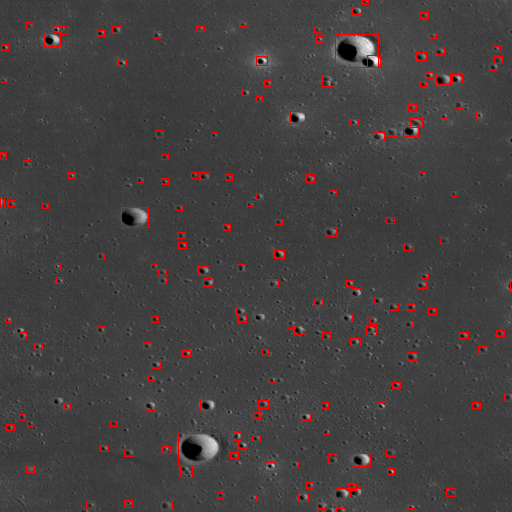}}
    \\[2mm]
    \subfloat[YOLOv26m]{\includegraphics[width=0.3\textwidth]{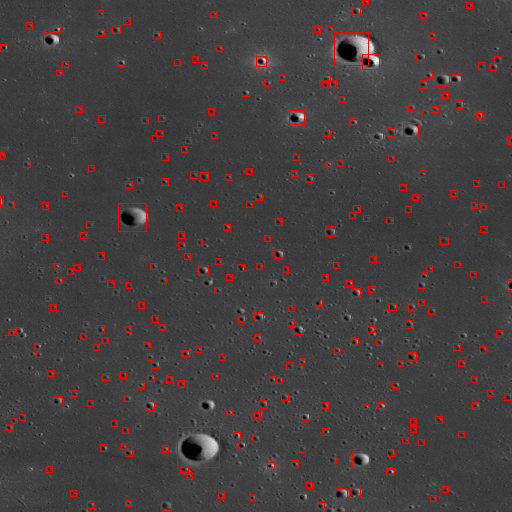}}\hspace{2mm}
    \subfloat[RT-DETR]{\includegraphics[width=0.3\textwidth]{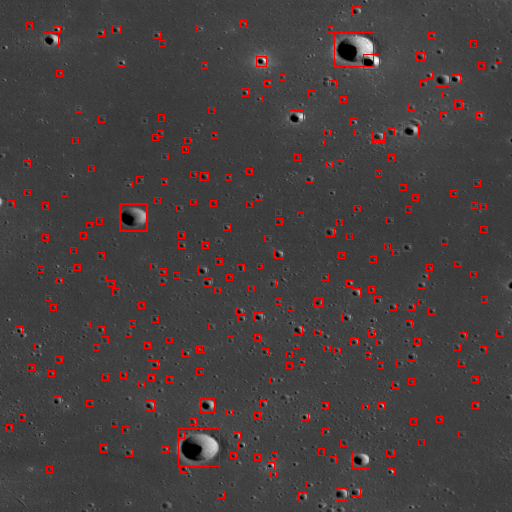}}\hspace{2mm}
    \subfloat[fine-tuned OWLv2 model]{\includegraphics[width=0.3\textwidth]{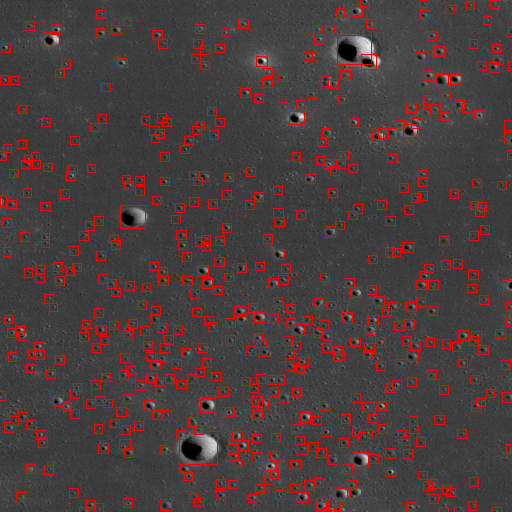}}
\caption{Qualitative comparison of detection methods on an out-of-distribution (OOD) tile of size 512$\times$512 from the LRO-WAC global morphologic map: (a) Original input tile. (b) OWLv2 zero-shot inference. (c) Few-shot approach introduced by Bauer et al.~\cite{Bauer2025}. (d) YOLOv26m results. (e) RT-DETR results. (f) Our proposed fine-tuned OWL model demonstrates the most accurate detection on this challenging OOD sample.}\label{fig:sotacomparison}
\end{figure*}

Table~\ref{tab:sota_comparison} shows our evaluation of the proposed method compared to the SoTA models. We compute the average precision and the average recall over the full test dataset of IMPACT. The results indicate that the fine-tuned OWLv2 model has significantly the highest recall value with 86.2$\%$ followed by RT-DETR with 77.0$\%$. Notably, all the YOLO models and RT-DETR have higher precisions; for example, YOLOv8n has a precision of 95.0$\%$. While the YOLO models have high precision, the recall values are rather low with YOLOv8n achieving 36.7$\%$, YOLOv11m 48.6$\%$ and YOLOv26m 55.6$\%$. This indicates that a substantial portion of ground truth craters is not detected by these models. As mentioned above, a lower precision value may be due to the detection of new craters rather than false positives because the underlying dataset contains many instances of missed annotations. Nevertheless, we emphasize that fine-tuning various versions of YOLO and RT-DETR yields very promising results. OWLv2's zero-shot capability has the lowest recall value by far, as it detects only a small fraction of craters and is not suitable for detecting craters when applied unmodified. However, by applying the few-shot approach with OWLv2~\cite{Bauer2025}, we note that it has the third highest recall value, emphasizing its capability as a crater detector, which is also supported by a solid precision score.

\begin{table}[!ht]
\centering
\caption{Quantitative comparison with other state-of-the-art models on the IMPACT test set. The Recall and Precision columns show the average values for the entire test dataset.}\label{tab:sota_comparison}
\begin{tabular}{l|cc}
\hline
Method & avg. Recall (\%) & avg. Precision (\%) \\
\hline
OWLv2 fine-tuned & \textbf{86.2} & 51.3 \\
OWLv2 zero-shot & 0.30 & 90.8 \\
OWLv2 few-shot  & 56.6 & 76.7 \\
YOLOv8n & 36.7 & \textbf{95.0} \\
YOLOv11m & 48.6 & 92.2 \\
YOLOv26m & 55.6 & 87.0 \\
RT-DETR & 77.0 & 59.8 \\
\hline
\end{tabular}
\end{table}

Fig.~\ref{fig:sotacomparison} further illustrates the effectiveness of the fine-tuning strategy of OWLv2. We select a tile of size 512$\times$512 of the LROC WAC global morphologic map (\url{https://data.lroc.im-ldi.com/lroc/view_rdr/WAC_GLOBAL}), which is a cylindrical projection with a scale of 100 m/px. This is an OOD image, as the background and the craters appear different from those in the training samples. We note that the zero-shot approach detects almost no craters due to the large domain gap. The few-shot approach presented in~\cite{Bauer2025} significantly improves detection quality. Both YOLOv26m and RT-DETR achieve remarkable results, while our proposed model detects the most craters across all approaches, suggesting superior generalization capabilities.

\section{DISCUSSION AND FUTURE WORK}

In this paper, we introduced a novel crater detection method based on fine-tuning the pre-trained OWLv2 model and an adapted loss function. This method aligns text and images in a shared embedding space using a text Transformer and a ViT, respectively. To do so, we inserted trainable parameters based on LoRA into the vision encoder as well as in classification and box head, respectively, while keeping the text Transformer frozen. An ablation study showed that training solely the ViT and training the ViT and the detection heads with varying LoRA ranks yield similar results, with the ViT-only configuration performing slightly better. We hypothesize that this may be due to the rather small dataset. Future work will incorporate a large amount of pseudo-labeled data, and we will further investigate how parameter injection in the heads may improve the results. We removed the objectness head entirely, as it is of no use during inference and would add unnecessary memory and computational overhead. For training and evaluation, we utilized a manually labeled dataset from the IMPACT project. During training, we aligned the predicted bounding boxes with the GT annotations by minimizing a CIoU-based loss, while simultaneously aligning the corresponding class vector by maximizing the cosine similarity with the anchor vector obtained from the text Transformer encoding the word ``crater''. 

Our method is qualitatively evaluated on randomly selected CDR images and quantitatively on the test dataset from IMPACT. The qualitative evaluation shows that the method successfully detects craters of different sizes and shapes under varying illumination conditions. Also, the model generalizes well, as demonstrated by its ability to detect more elliptical craters, such as those on lunar south pole images.  Visually, only a few false negatives occur, indicating that our model can be broadly applied to detect craters under various image conditions. This is supported by the quantitative evaluation, where recall values remain high, ranging from $80.3\%$ to $92.6\%$.
On the other hand, we observe notably lower precision values ranging from $28.0\%$ to $71.4\%$. We emphasize that many of the false positives affecting the precision metric are in fact  actual craters that were not labeled in the GT, either due to the accuracy score threshold or the minimum size constraint of 8 pixels. Nevertheless, we note that some detections are incorrectly classified as craters, especially in rugged lunar terrain. Further, the model appears to detect smaller craters more effectively than larger ones relative to the image size. This may be due to a learned bias towards smaller craters, given that the underlying IMPACT training data mostly comprises small craters. In future work, we will address this limitation by utilizing the fine-tuned model in a more advanced training strategy. This will involve using a semi-supervised learning approach with IMPACT data and pseudo labels. Although the proposed ViT-based crater detection model has good generalization capabilities, it is limited when it comes to off-nadir images. Although the craters are correctly identified in these images, the corresponding bounding boxes do not accurately enclose them. As OWLv2 is a very large model, it takes longer to perform inference than e.g.~a YOLO-based model. Therefore, it can only be used for ground observations and not for real-time on-board detection models that support navigation. A student-teacher based learning approach could address this challenge. Overall, we conclude, that applying a VLM to the task of crater detection yield promising results.

Future work will focus on improving the detection of the smallest craters and on extending the application to other celestial bodies, such as Mars. Further we also aim to respect the circular morphology of craters and extract their shapes directly rather than relying on bounding boxes, which could involve the use of semantic segmentation techniques. 
\section*{Acknowledgments}
This research work is conducted via the Open Space Innovation Platform (\href{https://ideas.esa.int}{https://ideas.esa.int}) as a Co-Sponsored Research Agreement and carried out under the Discovery programme of, and funded by, the European Space Agency (contract number: 4000144734). The view expressed in this publication can in no way be taken to reflect the official opinion of the European Space Agency. This work is also partially funded by the Hochschule Darmstadt and the University of Technology of Troyes within the European University of Technology EUT+.

\bibliographystyle{unsrturl}
\bibliography{references}

\end{document}